\documentclass[runningheads]{llncs}

\usepackage[T1]{fontenc}
\def\doi#1{\href{https://doi.org/\detokenize{#1}}{\url{https://doi.org/\detokenize{#1}}}}
\usepackage{xcolor} 
\usepackage{caption}
\usepackage{subcaption}

\usepackage{graphicx}
\usepackage{listings}
\begin{document}
\title{TraCon: A novel dataset for real-time traffic cones detection using deep learning}
%
%
\author{Iason Katsamenis \orcidID{0000-0001-9339-1546} \and
Eleni Eirini Karolou \orcidID{0000-0002-8546-3885} \and
Agapi Davradou \orcidID{0000-0002-6852-456X} \and
Eftychios Protopapadakis \orcidID{0000-0003-3876-0024} \and
Anastasios Doulamis \orcidID{0000-0002-0612-5889} \and
Nikolaos Doulamis \orcidID{0000-0002-4064-8990} \and
Dimitris Kalogeras \orcidID{0000-0002-9550-0088}}
\authorrunning{I. Katsamenis et al.}
%
\institute{National Technical University of Athens, 15780 Athens, Greece}

\maketitle              
\begin{abstract}
Substantial progress has been made in the field of object detection in road scenes. However, it is mainly focused on vehicles and pedestrians. To this end, we investigate traffic cone detection, an object category crucial for road effects and maintenance. In this work, the YOLOv5 algorithm is employed, in order to find a solution for the efficient and fast detection of traffic cones.  The YOLOv5 can achieve a high detection accuracy with the score of IoU up to 91.31\%. The proposed method is been applied to an RGB roadwork image dataset, collected from various sources.

\keywords{Traffic cones  \and Deep learning \and Computer vision \and Object detection.}
\end{abstract}
\section{Introduction}
Great changes have taken place in intelligent technology such as object detection in road networks \cite{protopapadakis2020multi}. Despite the technological progress, the demands for driving safety, efficiency, and automated maintenance systems have also increased significantly \cite{katsamenis2022simultaneous}. There are crucial challenges such as the capability to cope with temporary and sudden circumstances such as accidents and road construction \cite{dhall2019real}. Among the numerous objects, traffic cones are needed to be recognized since they present spatio-temporal visual appearance periodicity and are constantly replaced and moved in the road network. Traffic cone is an impermanent sign of traffic redirection, cordon-off an area, road accidents and lane shift \cite{yong2015real}.

The present paper outlines a deep learning approach to effectively recognizing traffic cones in roadwork images collected from multiple sources \cite{voulodimos2018deep}. This application was implemented with YOLOv5 algorithm which is widely used for object detection problems \cite{glenn_jocher_2020_3908560}, \cite{katsamenis2020man}. We created a dataset of RGB roadwork images that were annotated by engineer experts within the framework of the HERON project \cite{katsamenis2022robotic}. Traffic cone identification task can be addressed as an on-road object detection problem. The aim of our research is to broaden current studies of object detection issues and adapt them to the requirements of contemporary road network issues.

The aforementioned implementation can contribute to various studies that are related to traffic road efficiency and safety development. Firstly, it can support the observation of the pre/post-intervention phase including visual inspections in a roadwork project. Additionally, it can be beneficial in the intelligent transportation such as autonomous driving. In an automotive scenario, it can be used for avoiding obstacles while driving, predicting decisions and limitation of traffic accidents.

To this end, in this study we present a novel cone detection framework that includes: (i) formulating the identification as an object detection task, (ii) creating a roadwork image dataset, (iii) utilizing a YOLOv5 algorithm for object recognition.
The remainder of this paper is organized as follows. Section \ref{section:rel_work} briefly presents object detection frameworks that are related into object detection for road networks. Section \ref{section:proposed_system_arch} describes the proposed architecture. Section \ref{section:experimental_results} discusses the dataset description and the experimental results.

\section{Related work}
\label{section:rel_work}
The literature presents various noteworthy attempts at studies that use road images for object detection methods with deep learning techniques, in order to confront road issues \cite{katsamenis2020pixel}. Object detection methods can apply to different aspects of the above-mentioned issues. The work of \cite{perez2019deep} presents a single shot detection and classification of road users based on the real-time object detection system YOLO. This method is applied to the pre-processed radar range-Doppler-angle power spectrum. The study of \cite{kim2016road} suggests an on-road object detection using SSD which is a detection mechanism based on a deep neural network. In \cite{li2020detection} is proposed a novel deep learning anchor-free approach based on CenterNet for road object detection. The paper of \cite{pandey2018object} focuses on an object detection system called YOLO in order to enhance autonomous driving and other types of automation in transportation systems. Object detection is essential for automated driving and vehicle safety systems. For this purpose, the article \cite{haris2021road} compares five algorithms to inspect the contents of images, Region-based Fully Convolutional Network (R-FCN), Mask Region-based Convolutional Neural Networks (Mask R-CNN, Single Shot Multi-Box Detector (SSD), RetinaNet and YOLOv4.

Obstacle recognition on road images is another aspect of object detection. The work of \cite{sanil2020deep} implemented an obstacle detection and avoidance driverless car using Convolutional Neural Networks. In the paper of \cite{prabhakar2017obstacle} a deep learning system, using Faster Region-based convolutional neural network was employed for the detection and classification of on-road obstacles such as vehicles, pedestrians, and animals. Tsung-Ming Hsu et al. presented a deep learning model to mimic driving behaviors by learning the dynamic information of the vehicle along with image information in order to improve the performance of a self-driving vehicle. For the implementation of the model, they placed traffic cones on the road to collect the scene of avoiding obstacles \cite{hsu2018end}.

Little work has been presented in the literature on cone detection with deep learning techniques. The work of \cite{wang2020advanced} utilized a machine vision system with two monochrome cameras and two color cameras in order to recognize the color and position of traffic cones. Another approach is the study of \cite{arnold2019survey}, which presents an overview of object detection methods and used sensors and datasets in an autonomous driving application. \cite{seo2022temporary} focuses on the detection of a construction barrel, which includes a construction cone, a looper cone, a barricade, and four types of signs, via a collection of road images. Ankit Dhall et al. presented an accurate traffic cone detection and estimation of their position in the 3D world in real-time \cite{dhall2019real} presents an implementation of a robust autonomous driving algorithm using the Viola-Jones object detection method for traffic cones recognition. The study of \cite{albaranez2022case} proposes a lightweight neural network to perform cone detection from a racing car in order to research autonomous driving. Finally, the work of \cite{varghese2018changenet} presents a deep architecture called ChangeNet for detecting changes between pairs of images and expressing the same semantically. The dataset has 11 different classes of structural changes including traffic cones on road.

\section{Proposed System Architecture}
\label{section:proposed_system_arch}
The presented system utilizes the roadwork image dataset to identify traffic cones. Each image was properly fed into the YOLOv5 algorithm. YOLO is an acronym for 'You only look once' and is a target detection algorithm based on a regression algorithm that uses Neural Networks to provide real-time object detection. Its usefulness comes due to the fact that it completes the prediction of the classification and location information of the objects according to the calculation of the loss function, so it makes the target detection problem transform into a regression problem solution \cite{li2022two}. This algorithm extracts the most advanced detection technologies available at the time and optimizes the implementation for best practice \cite{ge2021yolox}. In this implementation, we utilize YOLOv5, which holds the best performance among YOLO algorithms. It is based on the PyTorch framework and its functionality comes from the fact that it is a suitable lightweight detector that can balance detection accuracy and model complexity under the constraints of processing platforms with limited memory and computation resources \cite{xu2022lite}.

\begin{figure}[h!]
\centering
\includegraphics[width=1\textwidth]{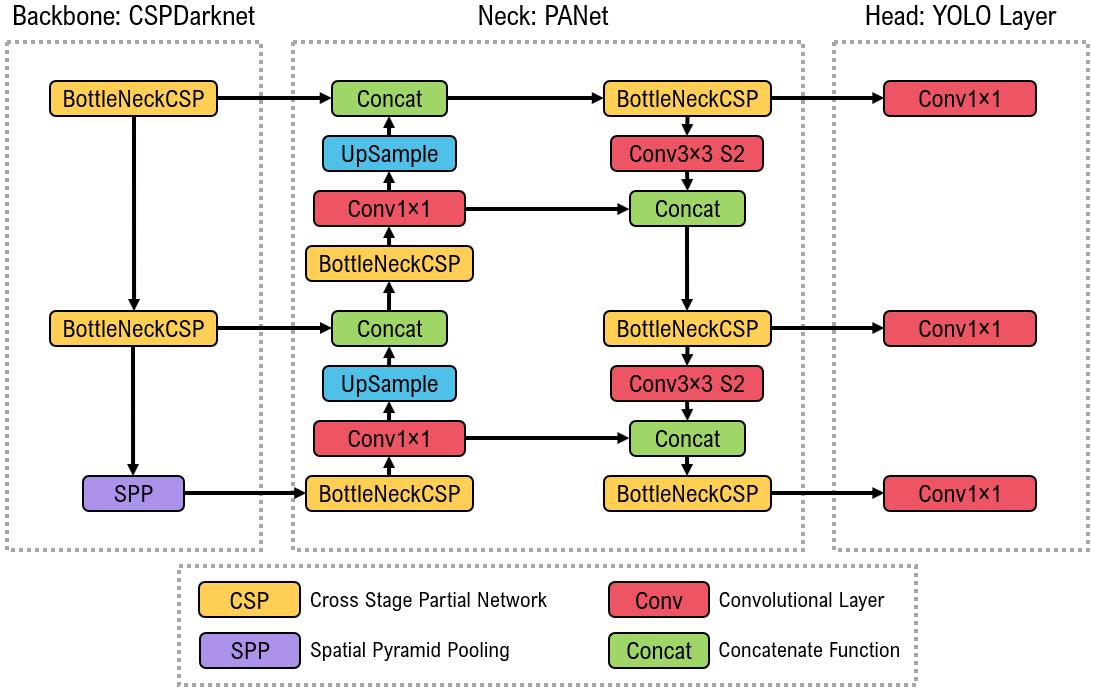}
\caption{The architecture of the YOLOv5 model, which consists of three parts: (i) Backbone: CSPDarknet, (ii) Neck: PANet, and (iii) Head: YOLO Layer. The data are initially input to CSPDarknet for feature extraction and subsequently fed to PANet for feature fusion. Lastly, the YOLO Layer outputs the object detection results (i.e., class, score, location, size).} 
\label{fig:yolo_arch}
\end{figure}

The architecture of the model YOLOv5 consists of three parts: (i) Backbone: CSPDarknet, (ii) Neck: PANet, and (iii) Head: YOLO Layer. The data are initially input to CSPDarknet for feature extraction and subsequently fed to PANet for feature fusion. Lastly, the YOLO Layer outputs the object detection results (i.e., class, score, location, size). The architecture of the model can be seen in Fig. \ref{fig:yolo_arch}.


\section{Experimental Evaluation}
\label{section:experimental_results}

\subsection{Dataset description}

The data used in this paper was collected and manually annotated under the framework of the H2020 HERON project \cite{katsamenis2022robotic}. HERON aims to develop an integrated automated system to perform maintenance and upgrading roadworks tasks, such as sealing cracks, patching potholes, asphalt rejuvenation, autonomous replacement of CUD (removable urban pavement) elements and painting road markings, but also supporting the pre- and post-intervention phase including visual inspections and dispensing and removing traffic cones in an automated and controlled manner.

\begin{figure}
     \centering
     \begin{subfigure}[b]{0.235\textwidth}
         \centering
         \includegraphics[width=\textwidth]{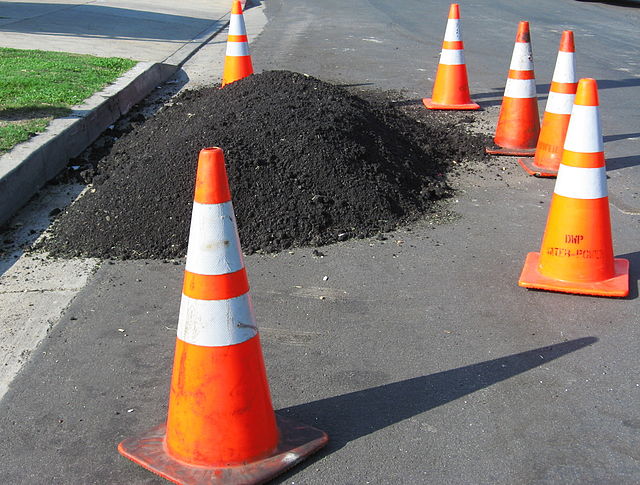}
         \caption{RGB image}
     \end{subfigure}
     \begin{subfigure}[b]{0.5\textwidth}
         \centering
         0 0.372656 0.093814 0.054688 0.183505\\
         0 0.705469 0.116495 0.092188 0.220619\\
         0 0.800000 0.183505 0.087500 0.280412\\
         0 0.856250 0.213402 0.090625 0.307216\\
         0 0.897656 0.387629 0.176563 0.453608\\
         0 0.347656 0.650515 0.329688 0.698969
         \caption{YOLO annotation format}
     \end{subfigure}
        \caption{Each RGB image (a) has a corresponding .txt file with the bounding box information (b) of the traffic cones (ID, x, y, w, h).}
        \label{fig:labelimg}
\end{figure}

More specifically, to train and evaluate the deep learning object detector, a dataset that contains RGB images was collected and manually annotated using labelImg \cite{LabelImg}, which is a graphical image annotation tool. labelImg is written in Python and uses Qt for its graphical interface. The produced annotations are saved as .txt files that store the information of the annotated bounding boxes. In particular, for each RGB image (see Fig. \ref{fig:labelimg}a) a corresponding text file was generated (see Fig. \ref{fig:labelimg}b) that contains a number of rows equal to the number of the bounding boxes (i.e., traffic cones) in the specific image. As one can observe in Fig. \ref{fig:labelimg}b, each row consists of five numbers: (i) An integer number, starting at 0, that represents the class ID, which therefore in our case always equals 0, since the cone detection task is a single class problem; (ii) the horizontal coordinate x of the central pixel of the bounding box; (iii) the vertical coordinate y of the central pixel of the bounding box;  (iv) the width w of the bounding box and (v) the height h of the bounding box. It is noted that the central position of the bounding box (ii-iii), as well its dimensions (iv-v) are real numbers on a scale of 0 to 1, and, therefore, represent the relative location and size of the bounding box with respect to the whole image.

The dataset contains RGB data from heterogeneous sources and sensors (e.g., DSLR cameras, smartphones, UAVs). Furthermore, the images vary in terms of illumination conditions (e.g., overexposure, underexposure), environmental landscapes (e.g., highways, bridges, cities, countrysides), and weather conditions (e.g., cold, hot, sunny, windy, cloudy, rainy, and snowy). In parallel, several images include various types of occlusions, thus making the traffic cone detection task more challenging.

\begin{figure}[h!]
\centering
\includegraphics[width=\textwidth]{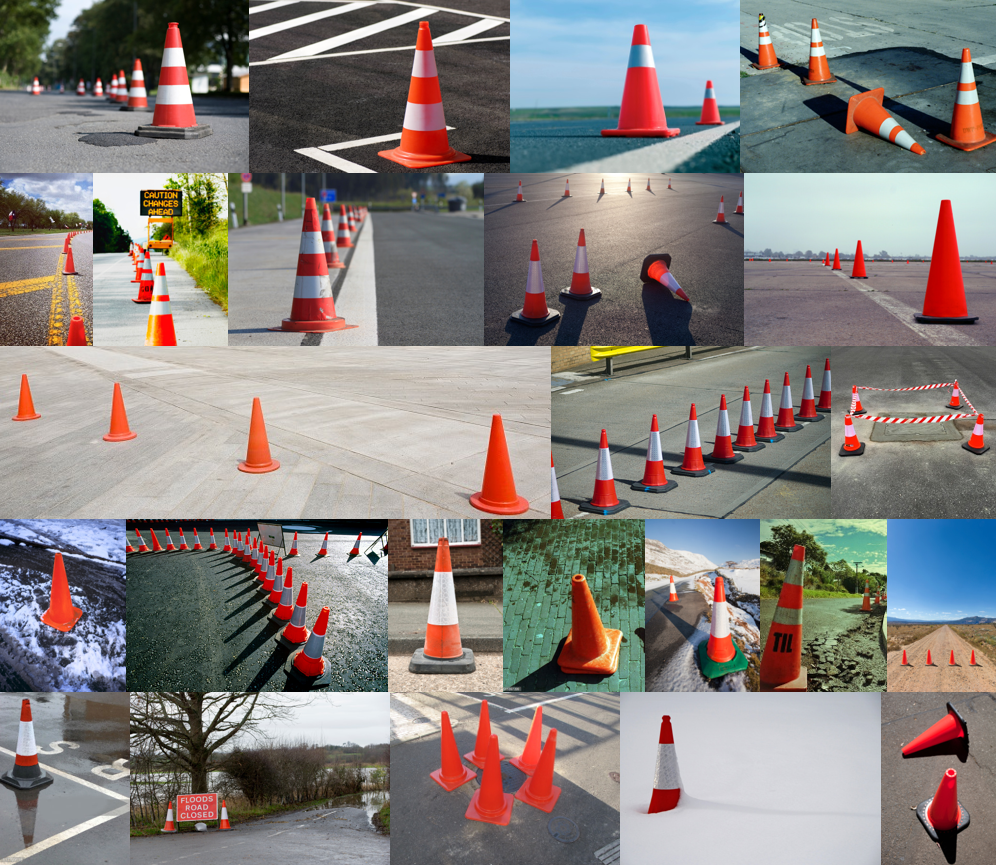}
\caption{Indicative images from the traffic cone detection dataset.} 
\label{fig:dataset}
\end{figure}

The total number of RGB images in the dataset is 540 with various resolutions ranging from 114×170 to 2,100×1,400. It is underlined that the total number of traffic cones in the entire dataset is 947. Representative samples of the dataset are demonstrated in Fig. \ref{fig:dataset}. From the images of the whole dataset, 92.5\% were used for training the deep model, and 7.5\% for testing its effectiveness. Among the training data, 80\% of them were used for training and the remaining 20\% for validation. The traffic cone detection dataset is made available online at: \url{https://github.com/ikatsamenis/Cone-Detection/} (accessed date 8 May 2022).

\subsection{Experimental setup - Model training}

Hence, for the training process, we utilized 500 images, 400 of which were included in the train set and 100 in the validation set. It is noted that the training data should include images with non-labeled objects (i.e., empty .txt files) and in particular, the negative samples without bounded boxes should be equal to the positive images with objects. To this end, 50\% of the data of both train and validation sets (i.e., 200 and 50 images respectively) are negative samples, while the rest contain at least one traffic cone. Lastly, it is underlined that to further generalize the learning process, we augmented the training data by horizontally flipping the corresponding images, thus increasing the train set size from 400 to 800.

The YOLO object detector was trained and evaluated using an NVIDIA Tesla K80 GPU with 12 GB of memory, provided by Google Colab. We trained the network, using batches of size 32, for 200 epochs, and set the input image resolution to 448×448 pixels. This work is based on the YOLOv5 small model in order to reduce the computational cost of the detection task. Towards this direction, the network takes up less than 15 MB of storage and thus can be easily embedded in smartphone applications and various low-memory digital devices or systems, including drones and microcontrollers.

\subsection{Evaluation metrics}

\begin{figure}[h]
\centering
\includegraphics[width=1\textwidth]{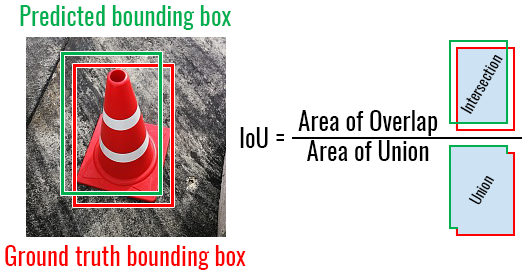}
\caption{Calculation of the IoU metric. The predicted bounding box is depicted in green color and the ground truth in red.} \label{IoU_metric}
\label{fig:iou_metric}
\end{figure}

The Intersection over Union (IoU) metric was employed in evaluating the performance of the proposed method. IoU is the most popular evaluation metric used in the object detection benchmarks \cite{rezatofighi2019generalized}. In order to apply IoU, ground-truth bounding boxes and predicted bounding boxes from our model are needed. This metric is used to evaluate how close the predicted bounding boxes are to the ground-truth bounding boxes. The greater the region of overlap, the greater the IoU, and therefore the detection accuracy as shown in Fig. \ref{fig:iou_metric}. Consequently, IoU is a number from 0 to 1 that specifies the size of the overlapping area between prediction and ground truth.

\subsection{Experimental Validation}
The proposed algorithm reached an excellent average IoU score of 91.31\%$\pm$5.42\% with a confidence level of 95\% over the data of the test set. Moreover, the network demonstrated an average prediction time of 0.065±0.029 seconds per image. 

\begin{figure*}[h]
    \centering
    \begin{subfigure}[h]{0.30\textwidth}
        \centering
        \includegraphics[width=1\linewidth]{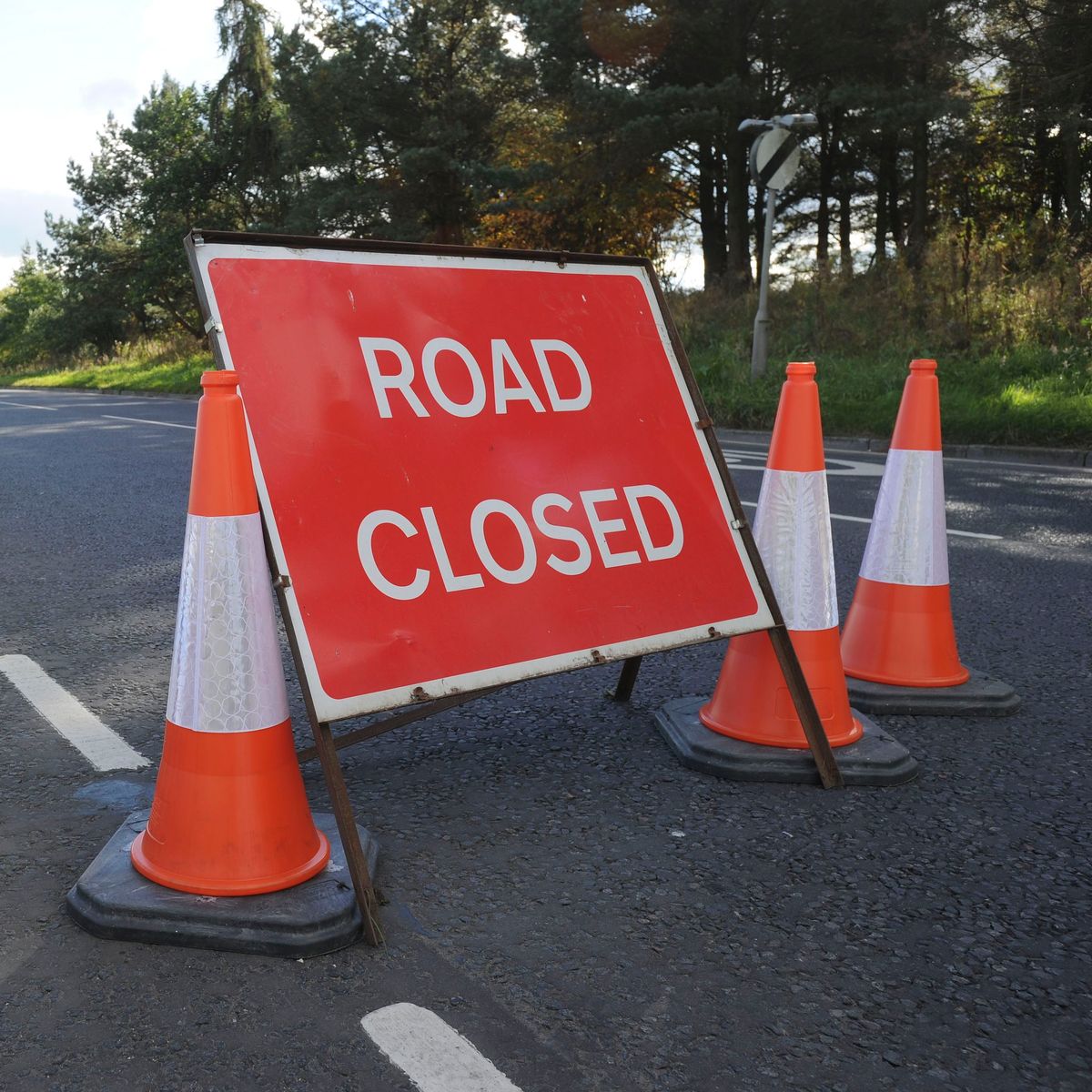}
    \end{subfigure}%
    ~ 
    \begin{subfigure}[h]{0.30\textwidth}
        \centering
        \includegraphics[width=1\linewidth]{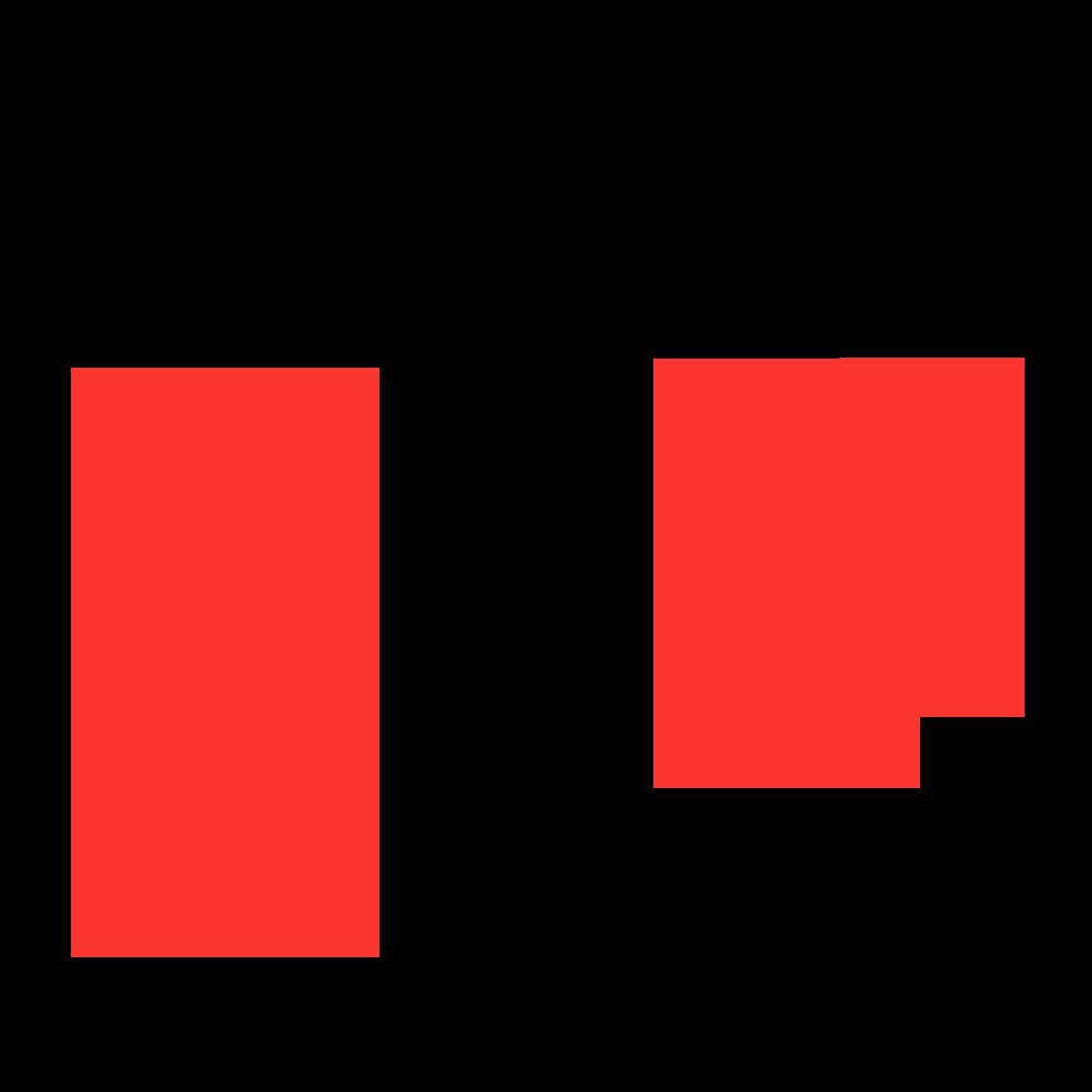}
    \end{subfigure}
    ~ 
    \begin{subfigure}[h]{0.30\textwidth}
        \centering
        \includegraphics[width=1\linewidth]{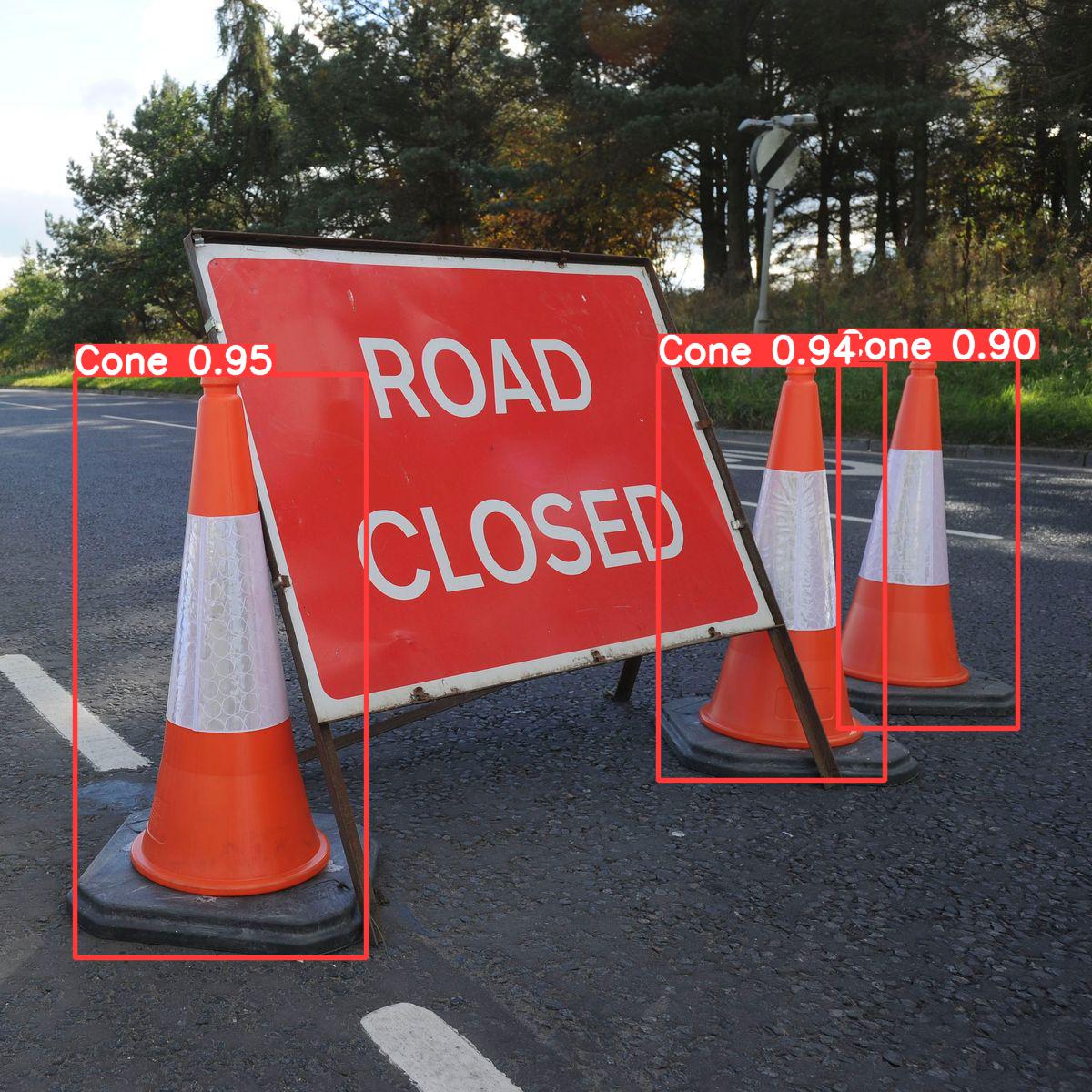}
    \end{subfigure}\\
        \centering
    \begin{subfigure}[h]{0.30\textwidth}
        \centering
        \includegraphics[width=1\linewidth]{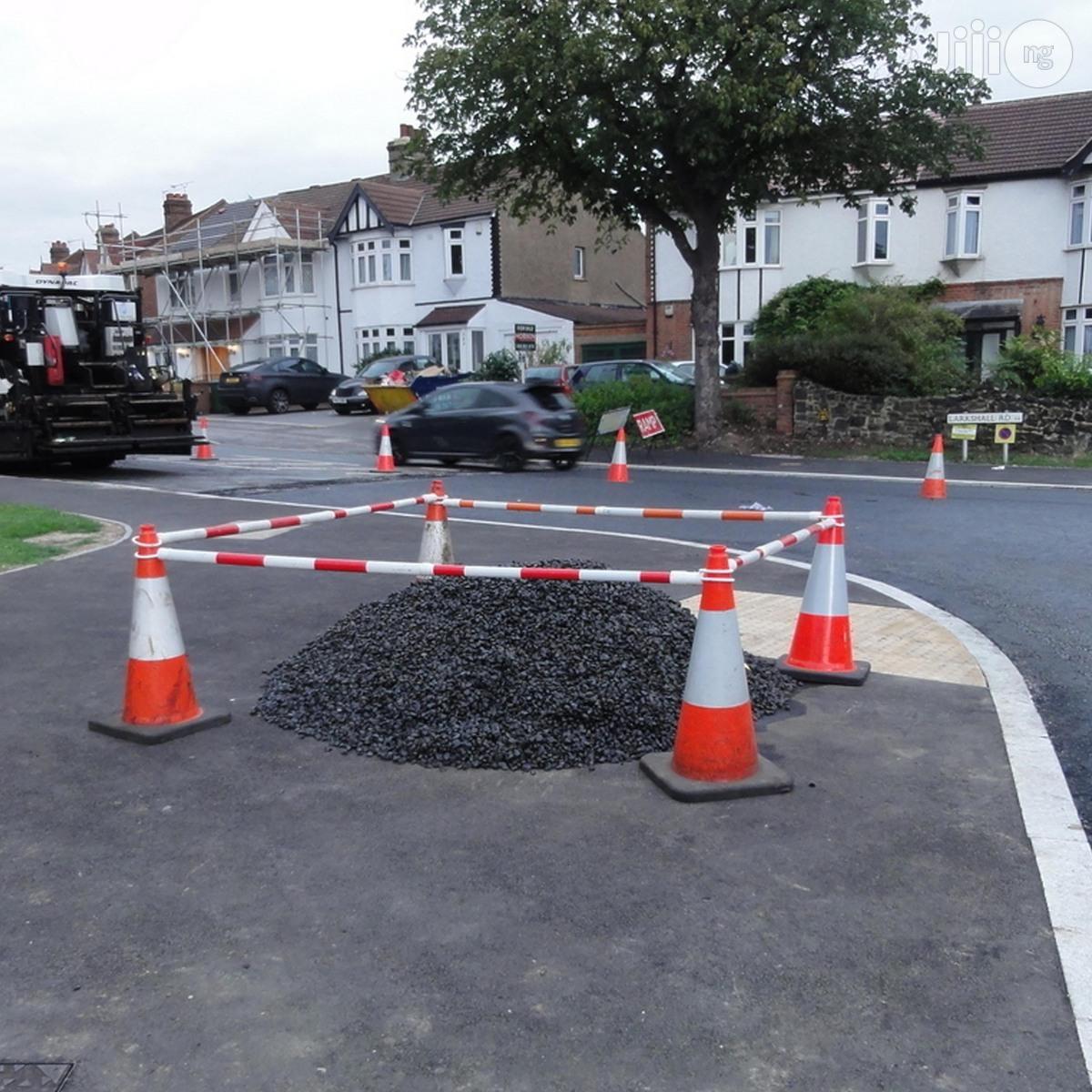}
    \end{subfigure}%
    ~ 
    \begin{subfigure}[h]{0.30\textwidth}
        \centering
        \includegraphics[width=1\linewidth]{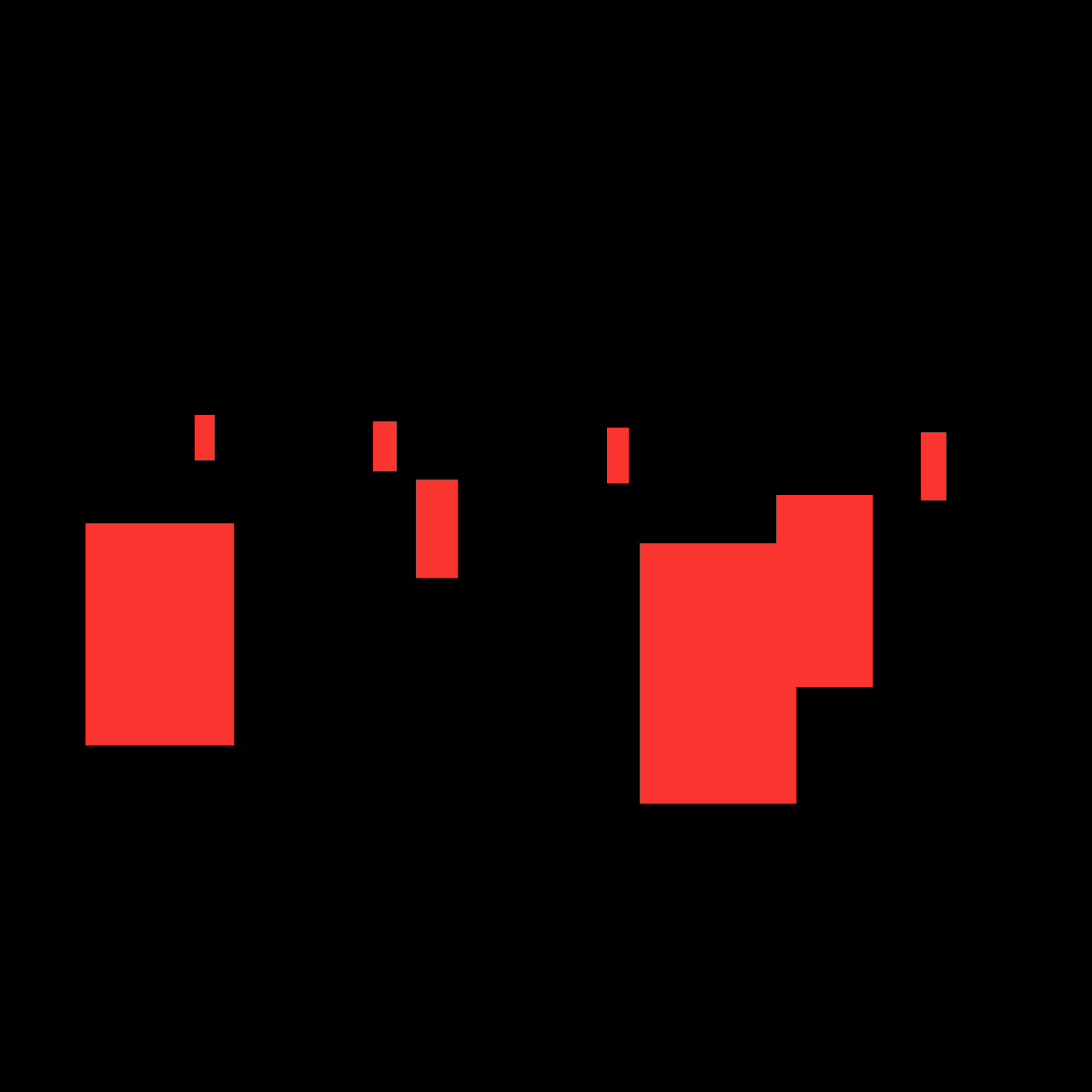}
    \end{subfigure}
    ~ 
    \begin{subfigure}[h]{0.30\textwidth}
        \centering
        \includegraphics[width=1\linewidth]{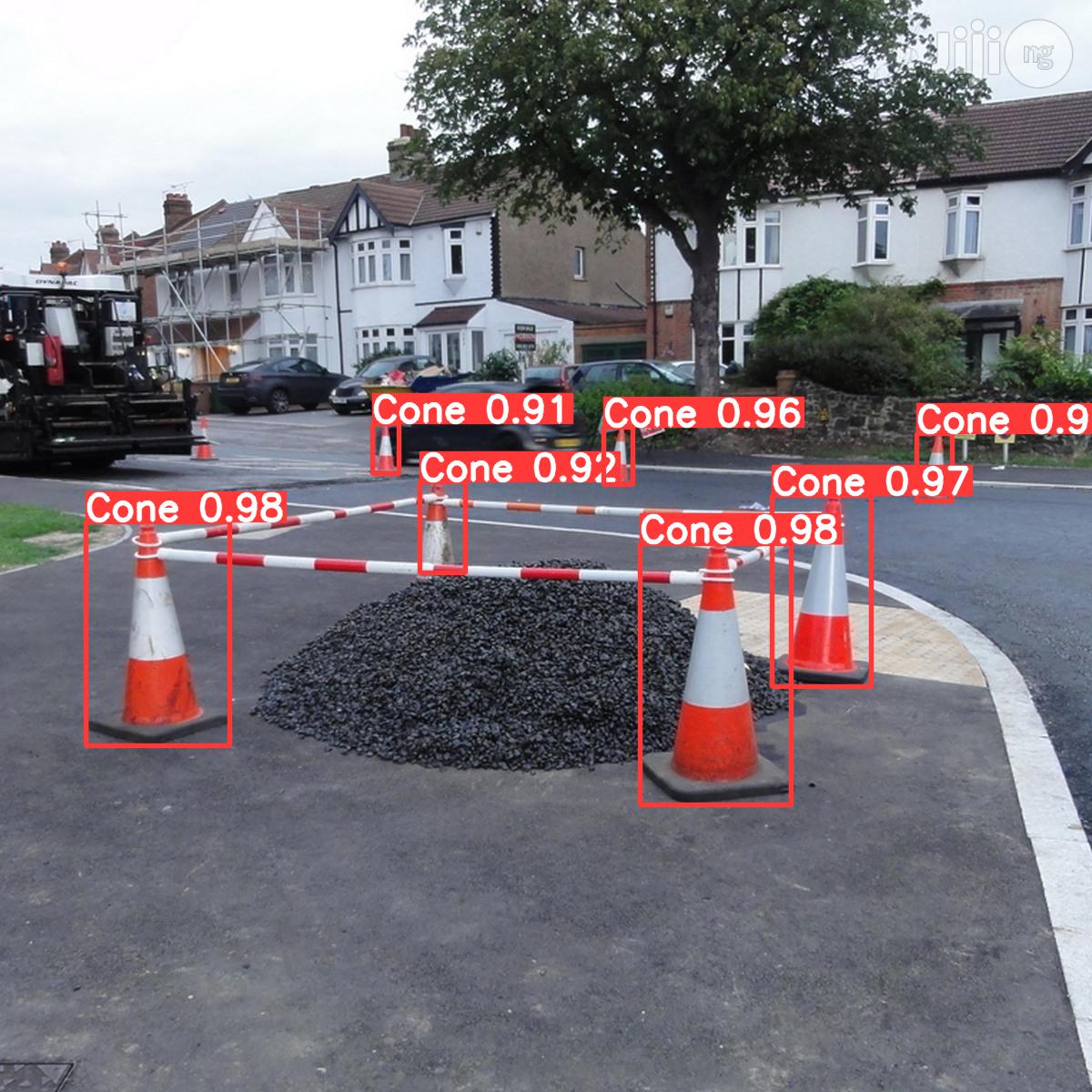}
    \end{subfigure}\\
        \centering
    \begin{subfigure}[h]{0.30\textwidth}
        \centering
        \includegraphics[width=1\linewidth]{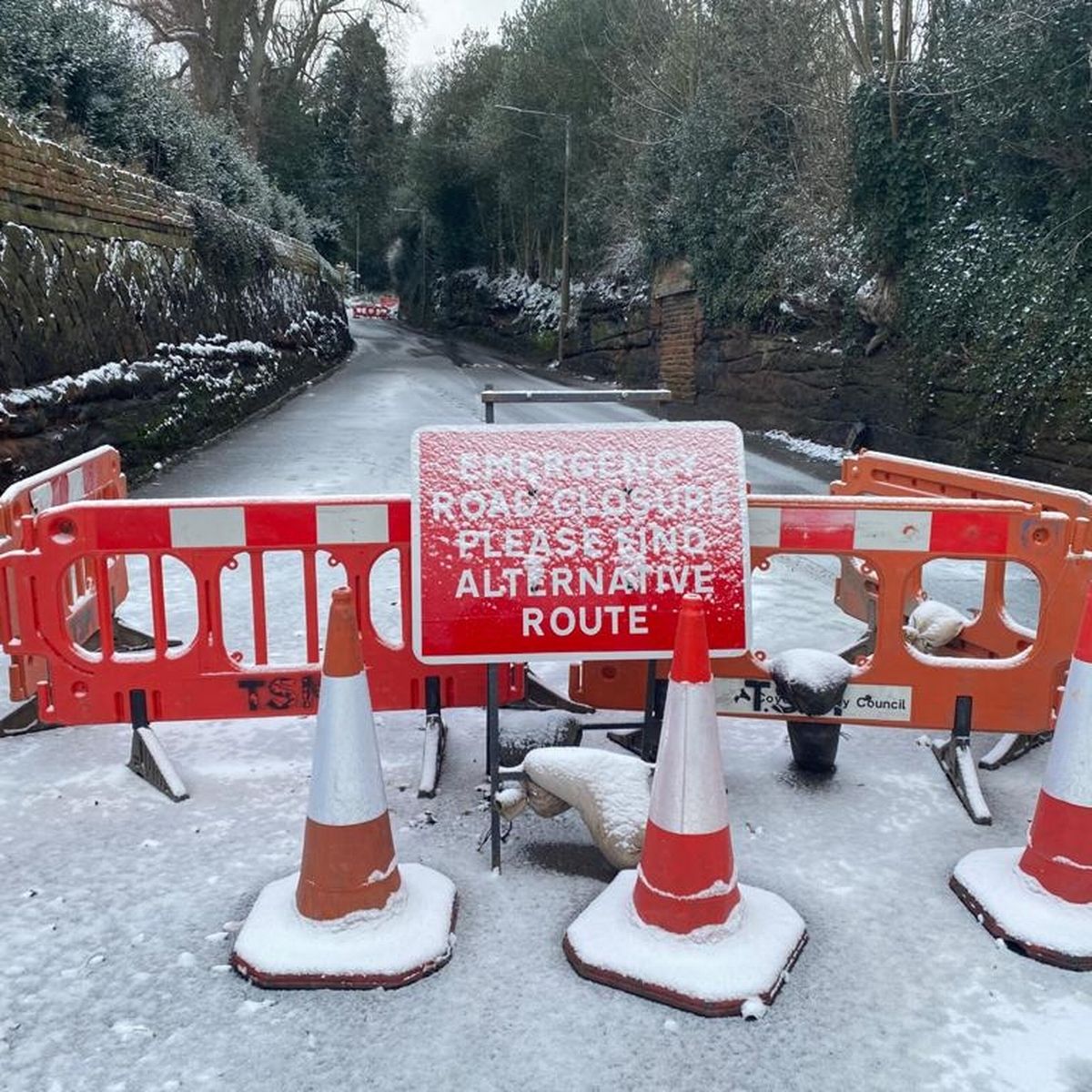}
        \caption{Original}
    \end{subfigure}%
    ~ 
    \begin{subfigure}[h]{0.30\textwidth}
        \centering
        \includegraphics[width=1\linewidth]{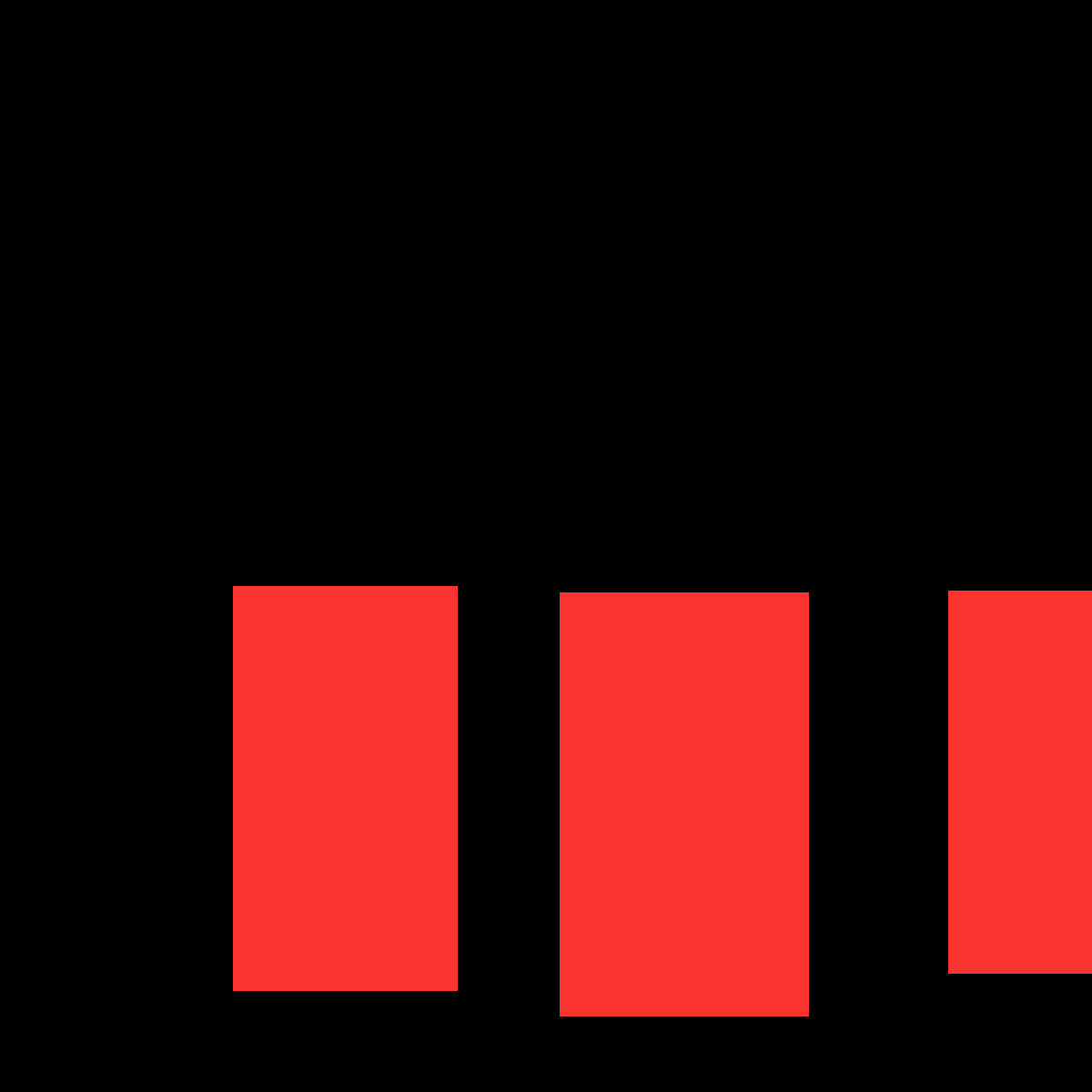}
        \caption{Annotation}
    \end{subfigure}
    ~ 
    \begin{subfigure}[h]{0.30\textwidth}
        \centering
        \includegraphics[width=1\linewidth]{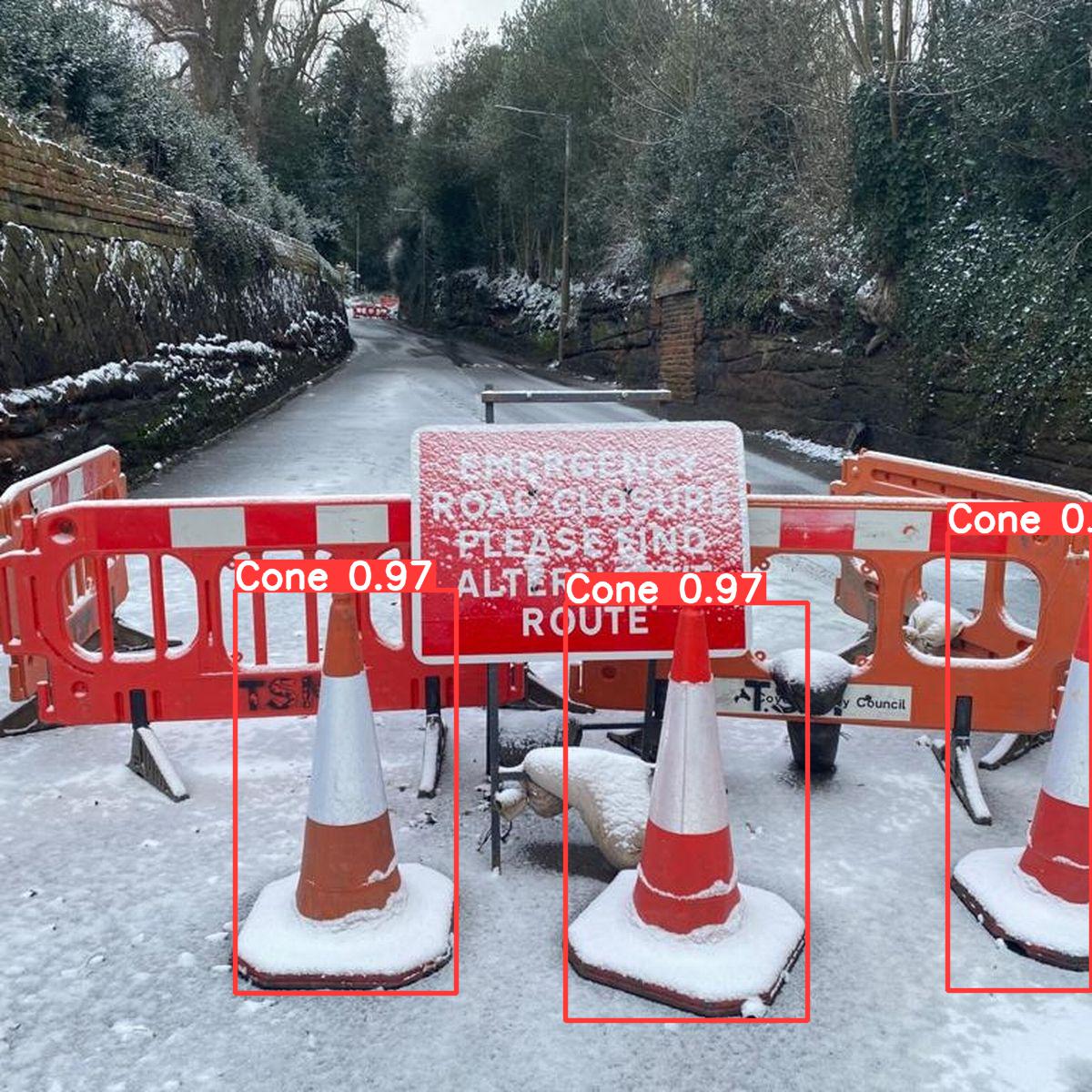}
        \caption{Prediction}
    \end{subfigure}
    \caption{Model's predictions (c) compared with the original (a) and annotated image (b).}
    \label{fig:yolo_predictions}
\end{figure*}

The experimental results using the YOLOv5 architecture are shown in Fig. \ref{fig:yolo_predictions}. The first column corresponds to the original RGB images followed by their ground truth bounding boxes in the second column. Finally, the last column illustrates the predicted bounding boxes with their corresponding confidence scores.

\section{Conclusions}

In this paper, we presented and evaluated a YOLOv5 algorithm for traffic cone recognition over a multisource roadwork image dataset. The utilized technique uses a deep learning framework, identifying traffic cones as an object detection scenario.  The model was able to achieve high scores and successfully managed the identification task. Future work should include the fusion of additional ephemeral objects that are correlated with road network maintenance and development.

\subsubsection{Acknowledgements} This work has received funding from the European Union’s Horizon 2020 Research and Innovation Programme under grant agreement No 955356 (Improved Robotic Platform to perform Maintenance and Upgrading Roadworks: The HERON Approach).

\bibliographystyle{splncs04}
\bibliography{bibliography}

\end{document}